\documentclass[3p,twocolumn]{elsarticle}
\makeatletter
\def\ps@pprintTitle{%
 \let\@oddhead\@empty
 \let\@evenhead\@empty
 \def\@oddfoot{\centerline{\thepage}}%
 \let\@evenfoot\@oddfoot}
\makeatother
\usepackage{amsmath,amssymb,amsfonts}
\usepackage{algorithmic}
\usepackage{graphicx}
\usepackage{textcomp}
\usepackage{wrapfig}
\usepackage{booktabs}
\usepackage[ruled,vlined]{algorithm2e}
\usepackage{algorithmic}
\usepackage{multirow}
\usepackage{subcaption}

\begin{document}

\begin{frontmatter}
\title{EaZy Learning: An Adaptive Variant of Ensemble Learning for Fingerprint Liveness Detection}
\author{Shivang Agarwal\fnref{aut1}}
\fntext[aut1]{shivanga.rs.cse16@iitbhu.ac.in}
\author{C. Ravindranath Chowdary \fnref{myfootnote}}
\fntext[myfootnote]{rchowdary.cse@iitbhu.ac.in}
\author{Vivek Sourabh\fnref{aut3}}
\fntext[aut3]{vivek.sourabh.cse14@iitbhu.ac.in}

\address{Department of Computer Science and Engineering, Indian Institute of Technology (BHU) Varanasi, India - 221005}
\begin{abstract}
In the field of biometrics, fingerprint recognition systems are vulnerable to presentation attacks made by artificially generated spoof fingerprints. Therefore, it is essential to perform liveness detection of a fingerprint before authenticating it. Fingerprint liveness detection mechanisms perform well under the within-dataset environment but fail miserably under cross-sensor (when tested on a fingerprint acquired by a new sensor) and cross-dataset (when trained on one dataset and tested on another) settings. To enhance the generalization abilities, robustness and the interoperability of the fingerprint spoof detectors, the learning models need to be adaptive towards the data.  We propose a generic model, EaZy learning which can be considered as an adaptive midway between eager and lazy learning. We show the usefulness of this adaptivity under cross-sensor and cross-dataset environments. EaZy learning examines the properties intrinsic to the dataset while generating a pool of hypotheses. EaZy learning is similar to ensemble learning as it generates an ensemble of base classifiers and integrates them to make a prediction. Still, it differs in the way it generates the base classifiers. EaZy learning develops an ensemble of entirely disjoint base classifiers which has a beneficial influence on the diversity of the underlying ensemble. Also, it integrates the predictions made by these base classifiers based on their performance on the validation data. Experiments conducted on the standard high dimensional datasets LivDet 2011, LivDet 2013 and LivDet 2015 prove the efficacy of the model under cross-dataset and cross-sensor environments. 
\end{abstract}
\end{frontmatter}

\section{Introduction}\label{sec1}
Fingerprint recognition systems are vulnerable to presentation attacks by numerous spoofing materials. As a countermeasure, fingerprint liveness detection plays a vital role in protecting the biometric systems by detecting the artificially reproduced fingerprints. Spoof fingerprint detectors determine the liveness of a fingerprint by classifying the fingerprint image into “live” or “spoof” category. The artificial replicas of “live” fingerprints can be easily generated using commonly available materials such as latex, wood glue, gelatin, silicon etc. 
The synthetic fingerprints generated using these materials are reported to have a success rate of more than $70\%$ when tested on various sensors \cite{Chugh0017}. Spoof detectors benefit fingerprint authentication systems in terms of increased security and user confidence \cite{eswa19,fgcs20}. Spoof detection becomes a challenging task when spoof fingerprints fabricated using new materials (that were not used in training) or sensed using unknown sensors are introduced to the detectors. Thus, making the fingerprint liveness detection an open set problem \cite{7180344}. Spoof detectors behave inadequately in the presence of fingerprints generated using fabrication materials that are currently unknown to the detectors \cite{Chugh019}.

Spoof detectors are majorly categorized into hardware-based and software-based mechanisms \cite{fgcs20}. In hardware-based mechanisms, a specialized hardware device monitors the additional life characteristics such as blood pressure, temperature, dry, moist or wet skin. 
Hardware-based solutions are useful in a static environment, but when the attacker finds a way to crack the system, it becomes difficult to upgrade the hardware device. Also, these devices are unprotected to attacks using new fabrication materials. 

Software-based spoof detection mechanisms rely on the representational features of fingerprint images to predict the class labels and are more appropriate for dynamic environments \cite{sensors19}. Various features have been used and proved to be efficient in training the spoof detectors for accurate classification. Still, the spoof detectors have a room for improvement in their performance while testing under cross-dataset and cross-sensor environments. The current state-of-the-art methods suffer from poor generalizability when tested under these environments.  

Therefore, the spoof detector must be robust to the changing environment and must perform reasonably well on cross-sensor and cross-dataset settings. 
\subsection{Learning Paradigms for Spoof Fingerprint Detection}
Spoof fingerprint detection is an application of biometrics and forensic science where the task is to classify fingerprint images into ``live” or ``spoof".  Ensemble learning has proved to be an adequate solution to this problem, but it does not provide adaptiveness towards the data. We claim that applications like spoof detection require the learning model to be adaptive to the properties intrinsic to the dataset. Therefore, we propose a novel learning scheme, EaZy learning which is midway between eager and lazy learning. 

Eager learning compiles the training data greedily and generates a concise hypothesis from the input samples and uses it for decision making. In contrast, lazy learning \cite{Aha1991} uses the input samples for decision making. Lazy learning is suited for applications where it is required to have good local approximations. Still, lazy learning requires to store the entire training data and defer the process of prediction until a query appears, which causes high memory consumption and low prediction efficiency. Thus making it challenging to use with practical applications.  Lazy learning incurs low computational cost during training but the high cost in responding to the queries. 

Therefore, we propose a novel learning model EaZy learning to overcome the challenges in learning paradigms. EaZy learning overcomes the high storage requirements and low prediction efficiency while maintaining good local approximations. The proposed model can be considered as a variant of ensemble learning which considers the properties of data and moves towards the eager or lazy nature of the learning paradigms.

EaZy learning differs from ensemble learning in the way it generates the ensemble and the way it integrates the outputs of the members of the ensemble. One of the major requirements of ensemble learning is to have a pool of diverse base classifiers \cite{1688199}. We achieve this by performing clustering on the training set and training the base classifiers on each cluster. In that way, we deliver diversity, which results in different generalization capabilities of base classifiers in the ensemble. EaZy learning is a plug-in solution capable of working with various base classifiers on any application. 

\subsection{Spoof Fingerprint Detection}
\begin{figure*}
	\centering
	\resizebox{\textwidth}{!}{\begin{tabular}{cccccc}
			\subcaptionbox{Live}{\includegraphics[width = 1in]{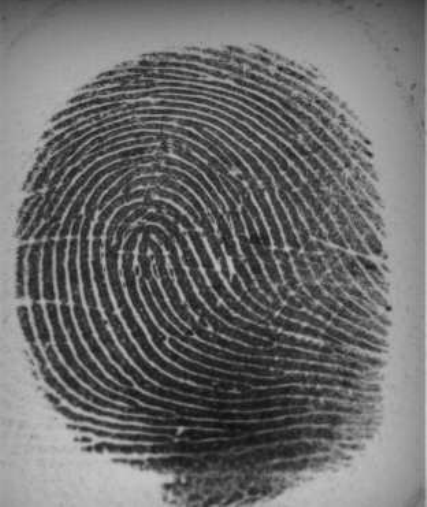}} &
			\subcaptionbox{Ecoflex}{\includegraphics[width = 1in]{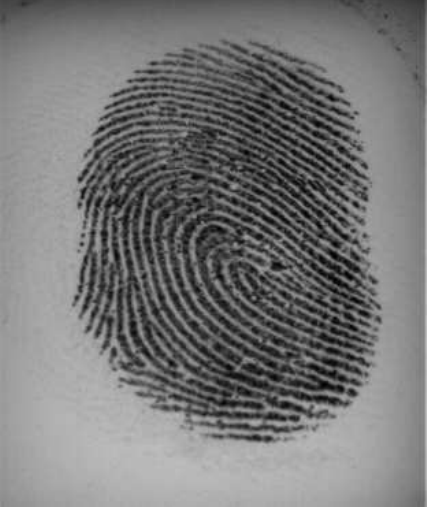}} &
			\subcaptionbox{Gelatin}{\includegraphics[width = 1in]{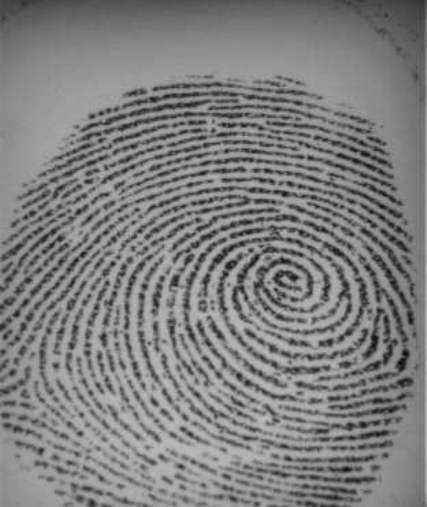}} &
			\subcaptionbox{Latex}{\includegraphics[width = 1in]{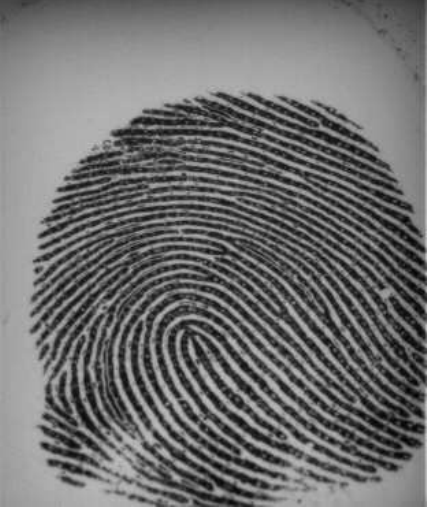}} &
			\subcaptionbox{Silgum}{\includegraphics[width = 1in]{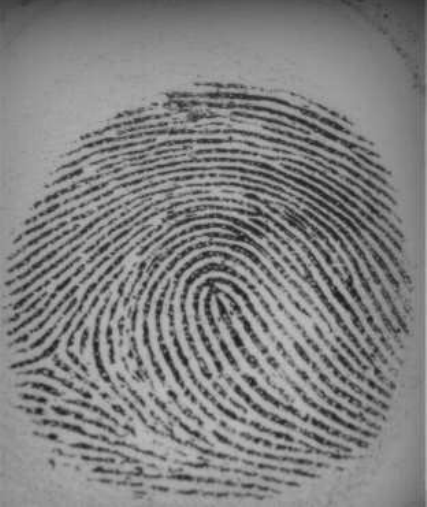}} &
			\subcaptionbox{WoodGlue}{\includegraphics[width = 1in, height = 1.2in]{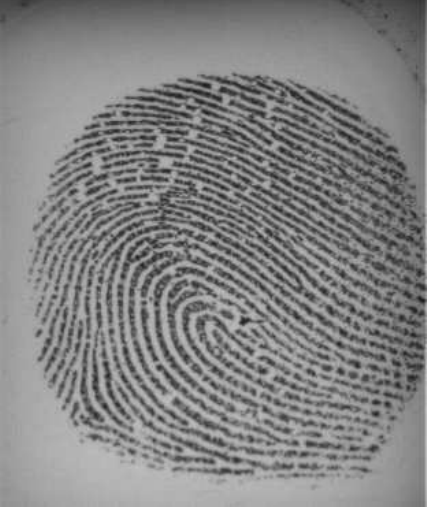}}\\
	\end{tabular}}
	\caption{Visual comparison between live and spoofs created using various spoof materials.}
	\label{fig:spoof}
\end{figure*}

In spoof fingerprint detection, the target is to prevent Presentation Attacks (PAs) caused by spoof fingerprints generated using various fabrication materials (Presentation Attack Instruments). A fingerprint authentication system requires to classify a fingerprint image into ``live" (bonafide presentation) or ``spoof" (attack presentation) before authenticating it. In general, spoofs sensed using new sensors (i.e., unknown to the learning model) may appear as test instances. Therefore, it is challenging for the spoof detector to keep on updating itself.  Figure \ref{fig:spoof} shows the visual comparison between live and spoof fingerprints. As can be observed, the human eye can not identify presentation attacks made on biometric systems.

Software-based solutions require extracting the features from the fingerprint images and classifying the image based on the learning model trained over those features \cite{7042825,7180344,AGARWAL2020113160,KHO201952,8306930}.
In this study, we propose an adaptive model which considers the spoof detection as a binary classification problem. In the past, the study of spoof fingerprint detection has targeted it as an application of closed-set supervised classification. We claim that cross-sensor and cross-dataset performances are of utmost importance, and the research in this field must try to tackle these difficulties. Therefore, we solve this problem by considering it as an open-set problem and make sure that the performance of the spoof detector is minimally compromised. 
The contributions made by this study are as follows:
\begin{enumerate}
	\item We propose an adaptive learning model EaZy learning which generates an ensemble of diverse base classifiers.
	\item We evaluate the performance of various ensemble learning models and EaZy learning for spoof fingerprint detection under cross-sensor and cross-datasets environments.
	\item We emphasize on adapting to the properties of data while generating the hypotheses and show robustness against the fingerprints generated using unknown fingerprint sensors.
\end{enumerate}

\section{Related Work}
The field of spoof fingerprint detection has evolved rapidly over recent years, and yet it demands more attention. Fingerprint recognition systems are vulnerable to Presentation Attacks (PAs) made by various materials. In an adversarial environment, the attacker keeps on evolving and attempts to break the system using a fabrication material on which the spoof detector has not been trained yet or by using a new sensor. Therefore, spoof fingerprint detection is an open-set problem, where the learning model needs to be adaptive towards the changing environment.   

In recent years, ensemble learning has been used extensively to come up with solutions for spoof detection. In \cite{KHO201952,AGARWAL2020113160}, the authors propose mechanisms based on ensemble learning for tackling with presentation attacks on fingerprint recognition systems. The robustness provided by multiple classifier systems is useful in many problems. Spoof detectors can be benefitted by ensemble-based classifiers provided the base classifiers are diverse and accurate. 

In some attempts, the researchers use a single classifier to train on fingerprint images and classify the test images \cite{8306930,7180344,tcds19,apin20,icb19,sensors20}. Some single classifiers outperform the ensemble-based classifiers, but mostly they are useful when the problem is targeted as a closed set problem. 

Ensemble learning (EL) is useful in real-world problems with broad-spectrum. EL was primarily proposed to reduce the variance in the decision making systems. In an ideal scenario, EL expects the base classifiers to be diverse and accurate. Diversity among the classifiers is of utmost importance because of the robustness provided by it. Also, if each base classifier is accurate on the part of the entire problem, the ensemble is likely to cover the whole problem spectrum.  In the past, there has been some research on utilizing diversity and accuracy of individual classifiers to come up with the final decision \cite{TSAI201446}. The popular approaches based on EL are bagging, stacking, random subspace, random forest, etc. These approaches have been used for solving various real-world problems, but we claim that for an open-set problem, the learning model needs to be adaptive. 

In \cite{8664594}, the authors propose a soft voting ensemble technique to combine the predictions, which, similarly to our proposed work considers the individual performances while assigning weights to the classifiers. 

Recently, there have been some attempts by researchers to address the poor generalization abilities of spoof detectors. In \cite{wacv20}, the authors consider the poor generalization ability of the spoof detectors and propose an end-to-end patch level network to address the issue. In \cite{tifs20} and \cite{Chugh019}, spoof fingerprint generalization approaches based on CNN architecture are proposed.   

We claim that our proposed work is different from the above-mentioned works in the way it generates the ensemble of base classifiers. It considers the properties of data to decide the number of base classifiers to be integrated in the ensemble. Therefore, the individual classifiers are diverse and accurate, which by definition of ensemble learning is an ideal scenario. We claim that this feature of the proposed model makes it suitable for open set problems like spoof fingerprint detection.

\section{EaZy Learning}
Our proposed work EaZy Learning uses the same foundation as ensemble learning, i.e., instead of taking one expert's opinion, let several experts discuss and come up with a decision. In Multiple Classifiers Systems (MCSs), our goal usually is to generate a set $\Pi$ containing $n$ hypotheses that are accurate and diverse \cite{BANFIELD200549}. Therefore,
\begin{equation}
\Pi=\{\psi_1, .., \psi_n\}
\end{equation}\label{eq1} 
where $\Pi$ is an ensemble of base classifiers $\psi_i$, and $\psi$ is defined as,
\begin{equation}
\psi: H \times T \rightarrow Y
\end{equation}\label{eq2}
where $H$ is a hypothesis which operates on a set $T$ of instances and results in a class label belonging to a set $Y$. Ideally, ensemble learning is expected to generate a pool of classifiers $\Pi$, such that it exploits the unique competencies of each base classifier $\psi_i$.

The generated hypotheses should be consistent with the subset of data $D_{\psi_i}$ on which they are trained and disjoint from each other, such that:
\begin{equation}\label{eq3}
D_{\psi_1}\cup D_{\psi_2}...\cup D_{\psi_n}=D
\end{equation}
and,
\begin{equation}\label{eq4}
D_{\psi_i}\cap D_{\psi_j}=\phi
\end{equation} 
EaZy learning satisfies Equations \ref{eq3} and \ref{eq4} by performing clustering on the training data $D$ and training base classifier $\psi_i$ on each $D_i$. It accommodates the nature of the given data and works well irrespective of its similarity quotient. It is capable of overcoming the drawback of eager learning, i.e., poor local approximations, the drawback of lazy learning, i.e. inefficiency in the classification phase, and the drawback of ensemble learning, i.e. lack of adaptiveness towards the data. To achieve an adaptive midway by maintaining more than one hypothesis, we need to extract the common features of the examples and group them according to the similarity inherently present in them. 
\begin{algorithm}[]
	\caption{EaZy learning for training a classifier $\Pi$.}
	\label{alg:algorithm1}
	\textbf{Input}: Training dataset $D$, validation data $V$, classifier learning algorithm $K$, clustering algorithm $C$\\
	\textbf{Output}: A set of classifiers $\Pi$, a set $W$ containing the weights for the classifiers in $\Pi$
	\begin{algorithmic}[1] 
		\STATE $\{c_1, c_2, .., c_n\} \leftarrow C(D)$.\\\textit{Perform clustering on training set $D$.}
		\FOR{$i$ =1 to $n$}
		\STATE $\psi_i\leftarrow K(c_i)$.
		\STATE $Acc_i$ $\leftarrow$ Performance of $\psi_i$ on V
		\ENDFOR
		\STATE $\Pi \leftarrow \{\psi_1, \psi_2, .., \psi_n\}$
		\STATE $W \leftarrow \{w_1, w_2, .., w_n\}$
		\\\textit{Weights determined by Equation \ref{eq5} }
		\STATE \textbf{return} $\Pi$
		\STATE \textbf{return} $W$
	\end{algorithmic}
\end{algorithm}

\begin{figure}
	\includegraphics[scale=0.8]{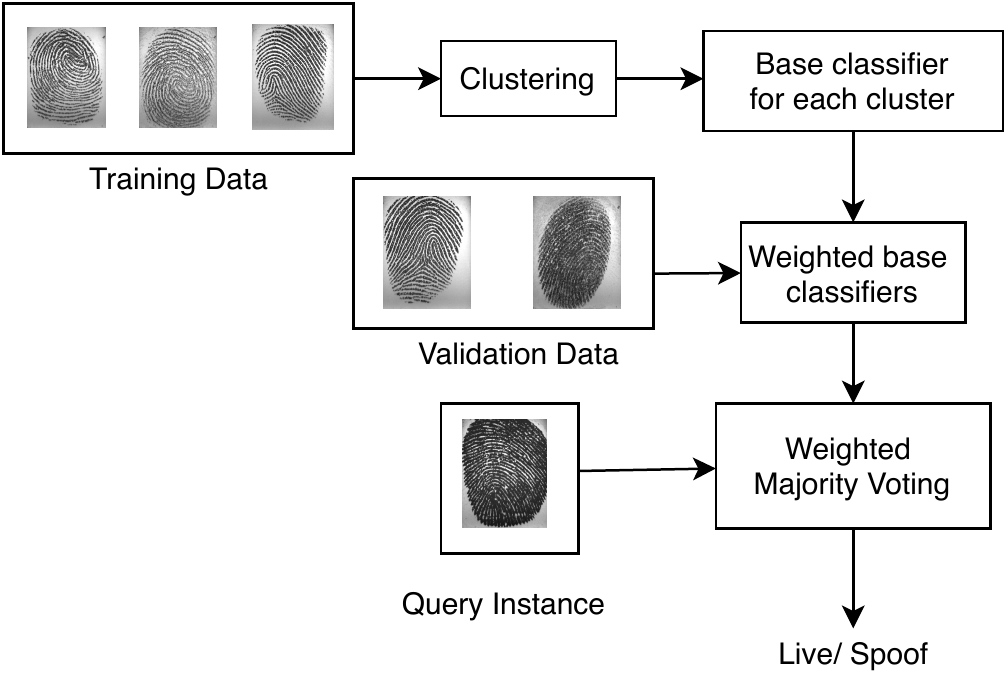}
	\caption{Conceptual Model of Adaptive Ensemble Learning. \cite{AGARWAL2020113160}}\label{fig:fig1}
\end{figure}

As represented in Figure \ref{fig:fig1}, we start with passing the training data into the training phase where data is partitioned into $n$ clusters based on the similarity present in the data. Later, we use base learners to generate hypotheses from these clusters. Each cluster $c_i$ yields one hypothesis $\psi_i$, which is trained only on records belonging to that cluster. The output of the training phase is $n$ hypotheses ($\psi_1, \psi_2, .., \psi_n$) that are accurate and diverse. The procedure of generating hypotheses for classification is given in Algorithm \ref{alg:algorithm1}. These hypotheses are given to the testing phase, where a query instance $x_q$ is assigned a discrete class by using weighted majority voting scheme defined by Equation \ref{eq5}.

Algorithm \ref{alg:algorithm1} is used for training a classifier $\Pi$. The inputs to the training phase are a training dataset $D$, a base classifier $K$ and a clustering algorithm $C$. We start by generating a validation set $V$ by randomly picking the instances from $D$. From our experiments, we found that the ideal size of $V$ is $20\%$ of $D$, but it may be changed depending on the application. 

Unlike Ensemble Learning, EaZy learning handles the training data in a different way. As explained in Figure \ref{fig:fig1} and Algorithm \ref{alg:algorithm1}, we generate a number of sub-datasets from the training data after performing clustering on it (Step 1 in Algorithm \ref{alg:algorithm1}). By performing clustering on the training data, we take the similarity inherently present in the dataset into consideration. Therefore, we do not require to define the number of classifiers apriori. In the best case, where we encounter a dataset containing all similar records, the proposed method converges itself to the eager learning paradigm and result in one hypothesis only. In the worst case, where we have a dataset containing dissimilar records, the proposed method converges itself to the lazy learning paradigm because the number of hypotheses is close to the number of training examples. On these generated sub-datasets, we use a base learning algorithm to generate the base classifiers or hypotheses (Step 3 in Algorithm \ref{alg:algorithm1}). These hypotheses are used for predicting the values for query instances. 

These classifiers are consistent with the data belonging to the respective cluster. The performance of these classifiers may not be equally good. Therefore, we use $V$ to test the performance of each classifier $\psi_i$ in $\Pi$ and learn the weightage $w_i$ for each $\psi_i$ as given in Equation \ref{eq5}. The validation set $V$ is passed to every $\psi_i$ to check its accuracy $A_i$. We determine the accuracy $A_i$ as a fraction of the number of instances correctly classified by $\psi_i$ to the total number of instances in $V$.

A high value of $A_i$ indicates the effectiveness of $\psi_i$. Therefore its weightage $W_i$ must be directly proportional to $A_i$. We use the Equation \ref{eq5} to determine the weightage of each $\psi_i$:

\begin{equation}\label{eq5}
W_i = \frac{A_i}{(\sum_{i}(A_i))} 
\end{equation}
We conclude the training phase of our model by generating a set of classifiers H with a set of weights W representing the weightage that should be given to the respective classifier while making a decision.
\section{Experimental Setup}\label{sec4}
The experimental setting is designed to demonstrate the working mechanism of EaZy learning and to explore the problem behaviour of spoof fingerprint detection under various environments. With our experiments, we aim to answer the following research questions:
\begin{itemize} 
	\item\textbf{RQ1:} How does EaZy learning perform under cross-sensor environment?
	\item\textbf{RQ2:} EaZy learning adapts to the properties of data while generating multiple base classifiers in the ensemble. How is it useful under cross-dataset environment?
\end{itemize} 
\subsection{Datasets}
The description of the datasets used in this study is given in Table \ref{dataset}. We use LivDet 2011 \cite{6199810}, LivDet 2013 \cite{6460190}, and LivDet2015 \cite{7358776} datasets used in fingerprint liveness detection competition held in subsequent years \cite{review19}. The goal of this competition is to compare software-based fingerprint liveness detection methodologies and fingerprint systems that are useful in identifying presentation attacks. Each of these datasets consists of live and spoof fingerprint images. These fingerprints are tested on biometric sensors such as Biometrika, DigitalPersona, ItalData, Sagem, CrossMatch etc. For each sensor, we have approximately 1000 fingerprint images belonging to the “live” category and the same number of images belonging to the “spoof” class. We have same number of images in training and testing. Further, the images belonging to the spoof class can be categorized in multiple sub-categories based on the fabrication material used for creating the spoof or fake fingerprint. These materials are gelatin, latex, playdoh, wood glue, silicone etc. The datasets have approximately 200 images, belonging to each of these sub-categories.
\begin{table*}	\caption{Description of datasets.}
	\centering
	\begin{tabular}{llll}
		\toprule
		Database   &                & Live (Train/Test) & Spoof (Train/Test)                                        \\ \midrule
		LivDet2011 & Biometrika     & 1000/1000        & 1000/1000 (ecoflex, gelatin, latex, silgum, wood glue)    \\
		& DigitalPersona & 1000/1000        & 1000/1000 (gelatin, latex, playdoh, silicone, wood glue)\\ 
		& ItalData       & 1000/1000        & 1000/1000 (ecoflex, gelatin, latex, silgum, wood glue)   \\
		& Sagem          & 1000/1000        & 1000/1000 (gelatin, latex, playdoh, silicone, wood glue) \\ 	
		LivDet2013 & Biometrika     & 1000/1000        & 1000/1000 (ecoflex, gelatin, latex, modasil, wood glue)   \\ 
		& ItalData       & 1000/1000        & 1000/1000 (ecoflex, gelatin, latex, modasil, wood glue)   \\ 
		& CrossMatch & 1250/1250 & 1000/1000 (body double, latex, playdoh, wood glue)\\
		LivDet2015 & Biometrika & 1000/1000 & 1000/1500 (ecoflex, gelatin, latex, RTV, wood glue)\\
		& DigitalPersona & 1000/1000 & 1000/1500 (ecoflex, gelatin, latex, RTV, wood glue)\\
		& CrossMatch & 1510/1500 & 1446/1448 (body double, ecoflex, gelatin, playdoh, oomoo)\\
		\bottomrule	
	\end{tabular}
	\label{dataset}
\end{table*}
\subsection{Features}
We use ResNet-50 model \cite{he2015deep} to extract the features from fingerprint images. ResNet-50 is a deep Residual Network originally designed for object recognition. ResNet-50 has been pre-trained on ImageNet database. By extracting the features using ResNet-50, we utilize transfer learning for spoof fingerprint detection. Due to space constraints, we refrain from discussing the ResNet-50 architecture in detail.
\subsection{Setup}
The motivation for proposing EaZy learning is to be able to generate an ensemble of base classifiers while considering the properties of data. Therefore, we use EM clustering algorithm \cite{Jin2010} to generate clusters of training instances. In that way, we get a pool of $n$ disjoint base classifiers at the end of the training phase without defining $n$ apriori. We consider a validation set V which is a hold-out set of the original training data. In this study, we take $20\%$ of the original training data for validation. The remaining $80\%$ data is used as the actual training data. Therefore, we have three separate sets: test, train and validation. The number of clusters obtained in every experiment for EaZy learning is written in parentheses. Experiments were conducted 10 times and the average results are reported. 

For applications like spoof fingerprint detection where the false negatives have a huge cost, it is important to report the accuracy of the model along with Attack Presentation Classification Error Rate (APCER). APCER is defined as:
\textit{\begin{quote}
 “proportion of attack presentations using the same PAI (Presentation Attack Instruments, e.g., spoof material) species incorrectly classified as bona fide presentations in a specific scenario.”
\end{quote}}
The conventional ensemble algorithms used for comparison are listed below:
\begin{itemize}
	\item \textbf{Random Sub-Space Method:} Random Sub-Space Method (RSM) \cite{709601} is also called attribute bagging because of its nature of bootstrapping the attributes of a dataset to generate new sets. It is a popular ensemble learning approach which attempts to reduce the correlation between estimators in an ensemble by training them on random samples of features instead of the entire feature set. In this study, we use RSM along with SMO \cite{Platt1998} as the base classifier. 
	\item \textbf{Random Forest:} Random Forest (RF) \cite{Breiman2001} is a popular ensemble learning algorithm that operates by constructing a multitude of decision trees at training time and outputting the class that is the mode of the classes (classification) or mean prediction (regression) of the individual trees.
	\item \textbf{Ada-Boost:} Ada-Boost or Adaptive Boosting \cite{10.1007/3-540-59119-2_166} is also an ensemble learning algorithm that can be used in conjunction with many types of weak learners to boost their performance.
	\item \textbf{Bagging:} Bagging or Bootstrapped Aggregating \cite{Dietterichl2002-DIEEL} is an ensemble approach where the dataset is partitioned to create bootstrapped samples. Base classifiers are trained on these samples, and final prediction is made by majority voting. In this study, we use it with SMO base classifier. 
\end{itemize}

In this study, we have considered two experimental settings to explore the behaviour of the proposed model.

\begin{enumerate}
	\item{\textbf{Category-1: inter sensor, same material performance evaluation-\\}} In Category-1, we evaluate the ability of the spoof detector in the cross-sensor environment. Therefore, we train the model on images acquired from one sensor and test it on images belonging to another sensor. For example, the model is trained on Biometrika 2011 train dataset and tested on ItalData 2011 test dataset. Cross sensor setting evaluates the generalization ability of the model appropriately.
	\item{\textbf{Category-2: inter dataset, same sensor, same material performance evaluation-\\}} Category-2 is designed to evaluate the model's capability under cross-dataset environment. These experiments demonstrate the model's robustness against unknown data. Therefore, it is trained on LivDet 2011 and tested on LivDet 2013 and vice-versa.  
\end{enumerate}  

\section{Results and Discussion}
\begin{table*}\centering\caption{Performance evaluation of EaZy learning on Category-1.}\label{tab:cat-3}
	\begin{tabular}{c|c|c|c|c|c|c|c|c|c|c}
		\toprule
		\multirow{2}{*}{Dataset} & \multicolumn{2}{c|}{EaZy} & \multicolumn{2}{c|}{RSM(SMO)} & \multicolumn{2}{c|}{Bagging(SMO)} & \multicolumn{2}{c|}{AdaBoost} & \multicolumn{2}{c}{RF} \\ \cmidrule{2-11} 
		& Acc.          & APCER     & Acc.          & APCER         & Acc.            & APCER           & Acc.          & APCER         & Acc.       & APCER      \\ \midrule
		Bio-Ital2011 & 54.05(13) & 0.08  & 50.5     & 0.97  & 51.3         & 0.91  & 50.45    & 0.94  & 53.5     & 0.49  \\ \midrule
		Ital-Bio2011 & 50.5(10)  & 0.79  & 51.2     & 0.41  & 53.55        & 0.21  & 52.55    & 0.33  & 47.85    & 0.94  \\ \midrule
		Sag-Dig2011  & 50.55(9)  & 0.83  & 54.95    & 0.29  & 53.9         & 0.26  & 54.9     & 0.25  & 50.5     & 0.36  \\ \midrule
		Dig-Sag2011  & 67.96(16) & 0.47  & 57.35    & 0.03  & 56.66        & 0.02  & 56.81    & 0.04  & 62.01    & 0.32  \\ \midrule
		Bio-Ital2013 & 95.4(11)  & 0.06  & 83.1     & 0     & 82.75        & 0     & 72.6     & 0     & 74.65    & 0.48  \\ \midrule
		Ital-Bio2013 & 57.9(6)   & 0.83  & 93.05    & 0.03  & 86.75        & 0.02  & 91.6     & 0.02  & 71.05    & 0.43  \\ \midrule
		Bio-Dig2015  & 78.6(7)   & 0.19  & 73.16    & 0.2   & 73.72        & 0.16  & 72.48    & 0.21  & 72.8     & 0.26  \\ \midrule
		Dig-Bio2015  & 72.16(12) & 0.24  & 62.8     & 0.56  & 60           & 0.61  & 61.6     & 0.58  & 62.68    & 0.45  \\ \midrule
		Average      & 65.89     & 0.44  & 65.76    & 0.31  & 64.83        & 0.27  & 64.12    & 0.3   & 61.88    & 0.47  \\ \bottomrule
	\end{tabular}
\end{table*}
Next, we answer the research questions raised in Section \ref{sec4} on the basis of our experimental results:
\begin{table*}\begin{center}

\caption{Performance evaluation of EaZy learning on Category-2.}\label{tab:cat-4}
	\begin{tabular}{c|c|c|c|c|c|c|c|c|c|c}
		\toprule
		\multirow{2}{*}{Dataset} & \multicolumn{2}{c|}{EaZy} & \multicolumn{2}{c|}{RSM(SMO)} & \multicolumn{2}{c|}{Bagging(SMO)} & \multicolumn{2}{c|}{AdaBoost} & \multicolumn{2}{c}{RF} \\ \cmidrule{2-11} 
		& Acc.          & APCER     & Acc.          & APCER         & Acc.            & APCER           & Acc.          & APCER         & Acc.       & APCER      \\ \midrule
		Bio2011-13  & 67.1(10)  & 0.49  & 59.8     & 0.56  & 58.8         & 0.55  & 58.6     & 0.53  & 51.8     & 0.72  \\ \midrule
		Bio2013-11  & 60.7(11)  & 0.21  & 52.8     & 0.94  & 52.9         & 0.94  & 53.35    & 0.93  & 53.85    & 0.91  \\ \midrule
		Ital2011-13 & 80.55(5)  & 0.04  & 67.9     & 0.63  & 67.4         & 0.63  & 68.65    & 0.6   & 63.8     & 0.59  \\ \midrule
		Ital2013-11 & 51.15(9)  & 0.04  & 51.15    & 0.98  & 51.15        & 0.98  & 51.4     & 0.97  & 54.35    & 0.89  \\ \midrule
		Bio2013-15  & 72.68(12) & 0.18  & 63.8     & 0.46  & 64.12        & 0.44  & 69.36    & 0.17  & 40.88    & 0.98  \\ \midrule
		Bio11-15    & 46.48(17) & 0.03  & 45.16    & 0.91  & 46.56        & 0.88  & 48       & 0.85  & 40.42    & 0.93  \\ \midrule
		Bio2015-13  & 51.5(14)  & 0.32  & 49       & 0.93  & 49.15        & 0.88  & 49.95    & 0.93  & 53.85    & 0.68  \\ \midrule
		Bio2015-11  & 53.75(7)  & 0.15  & 56       & 0.5   & 53.85        & 0.62  & 52.15    & 0.49  & 48.6     & 0.66  \\ \midrule
		Average     & 60.49     & 0.18  & 55.7     & 0.74  & 55.49        & 0.74  & 56.43    & 0.68  & 50.94    & 0.79  \\ \bottomrule
	\end{tabular}
\end{center}
\end{table*}

As we mentioned earlier, the problem of fingerprint liveness detection must be projected as an open-set problem, where the test set may contain instances of unknown type, i.e., test-set may have attack presentations generated using various sensors that are not known to the training model. We claim that the proposed work EaZy learning is adaptive to the properties of the dataset. This adaptive nature is useful in cross-sensor and cross-dataset environment. As in real-world scenario, new sensors are used for authentication; a spoof detector needs to detect the new spoofs without requiring to be trained on those sensors. Table \ref{tab:cat-3} and Table \ref{tab:cat-4} demonstrate the algorithm's ability to adjust in cross-sensor and cross-dataset environment. In category-1, the performance of EaZy learning is better than its counterparts. The highest accuracy is $95.4\%$ on Bio-Ital 2013 datasets. Similarly, in category-2, EaZy learning outperforms its rivals with the highest accuracy of $80.55\%$ on ItalData 2011-13 datasets. 
\subsection{Discussion}
We positioned this study as an adaptive midway between eager and lazy learning. Based on its similarity with ensemble learning (EL), EaZy learning can also be considered as a different variant of Multiple Classifier Systems (MCS) paradigm. Following the motivation, we conducted our experiments under various settings. We focused on the adaptiveness towards the data while deciding the number of base classifiers constituting the ensemble. We emphasized that a spoof detector must learn from the similarity inherently present in the data. Therefore, for a spoof detector to be robust towards the presentation attacks made using unknown biometric sensors, adaptiveness plays an important role. We demonstrated that EaZy learning is best suited for this task. From Table \ref{tab:cat-3} and Table \ref{tab:cat-4}, it is evident that EaZy learning is the best choice for such environments. We performed Friedman significance test on category-2 APCER values. It is observed that EaZy learning performs significantly better than the rivals where $p$-value is $0.00112$ and ${\chi^2}_r$ value is 18.225. The significance level was set at $0.05$. Since $p$-value is less than the significance lever $\alpha$, the null hypothesis can be rejected. From these tests we can conclude that the proposed model is significantly better than the rival models under these environments.          
\section{Conclusions}
Fingerprint spoof detection is an open-set problem which requires the learning model to be adaptive. Spoof detectors need to identify presentation attacks made by using unknown presentation attack instruments or biometric sensors. We claim that a spoof detector must adapt to the features of the bonafide presentation (live fingerprint) and attack presentation (spoof) to perform reasonably well under cross-sensor and cross-dataset environments. EaZy learning is an adaptive model which proves to be robust under such extreme environments. We raised two research questions and demonstrated the experimental results answering these questions. We performed our experiments on standard high dimensional datasets and explored the working mechanism of the proposed model, along with the traditional ensemble learning models under various experimental settings. In future, we strive to use better image processing feature-sets to further improve the accuracy of classification.
\bibliographystyle{unsrt}
\bibliography{refer_ai.bib}
\end{document}